\documentclass[letterpaper]{article} 
\usepackage{aaai2026}
\usepackage{times}  
\usepackage{helvet}  
\usepackage{courier}  
\usepackage[hyphens]{url}  
\usepackage{graphicx} 
\usepackage{natbib}  
\usepackage{caption} 
\usepackage{algorithm}
\usepackage{algorithmic}
\usepackage{pifont}
\usepackage{amsmath}
\usepackage{booktabs} 
\usepackage{amssymb}

\usepackage{newfloat}
\usepackage{listings}
\DeclareCaptionStyle{ruled}{labelfont=normalfont,labelsep=colon,strut=off} 
\floatstyle{ruled}
\newfloat{listing}{tb}{lst}{}
\floatname{listing}{Listing}
\title{Rescind: Countering Image Misconduct in Biomedical Publications with Vision-Language and State-Space Modeling}

\author{
    Soumyaroop Nandi\textsuperscript{1,2}, Prem Natarajan\textsuperscript{1,2,3}
}

\affiliations{
    \textsuperscript{1}Information Sciences Institute, University of Southern California, Marina Del Rey, CA, USA\\
    \textsuperscript{2}Thomas Lord Department of Computer Science, University of Southern California, Los Angeles, CA, USA\\
    \textsuperscript{3}Capital One, USA\\[3pt]
    soumyarn@usc.edu, premkumn@usc.edu
}

\usepackage{bibentry}

\begin{document}

\maketitle

\begin{abstract}
Scientific image manipulation in biomedical publications poses a growing threat to research integrity and reproducibility. Unlike natural image forensics, biomedical forgery detection is uniquely challenging due to domain-specific artifacts, complex textures, and unstructured figure layouts. We present the first vision-language guided framework for both generating and detecting biomedical image forgeries. By combining diffusion-based synthesis with vision-language prompting, our method enables realistic and semantically controlled manipulations—including duplication, splicing, and region removal—across diverse biomedical modalities. We introduce Rescind, a large-scale benchmark featuring fine-grained annotations and modality-specific splits, and propose Integscan, a structured state-space modeling framework that integrates attention-enhanced visual encoding with prompt-conditioned semantic alignment for precise forgery localization. To ensure semantic fidelity, we incorporate a VLM-based verification loop that filters generated forgeries based on consistency with intended prompts. Extensive experiments on Rescind and existing benchmarks demonstrate that Integscan achieves state-of-the-art performance in both detection and localization, establishing a strong foundation for automated scientific integrity analysis.

\end{abstract}

\begin{links}
    \link{Code}{https://github.com/SoumyaroopNandi/Rescind}
\end{links}

\section{Introduction}
\label{sec:intro}

\begin{figure}[t]
    \centering
    \includegraphics[width=1.02\linewidth]{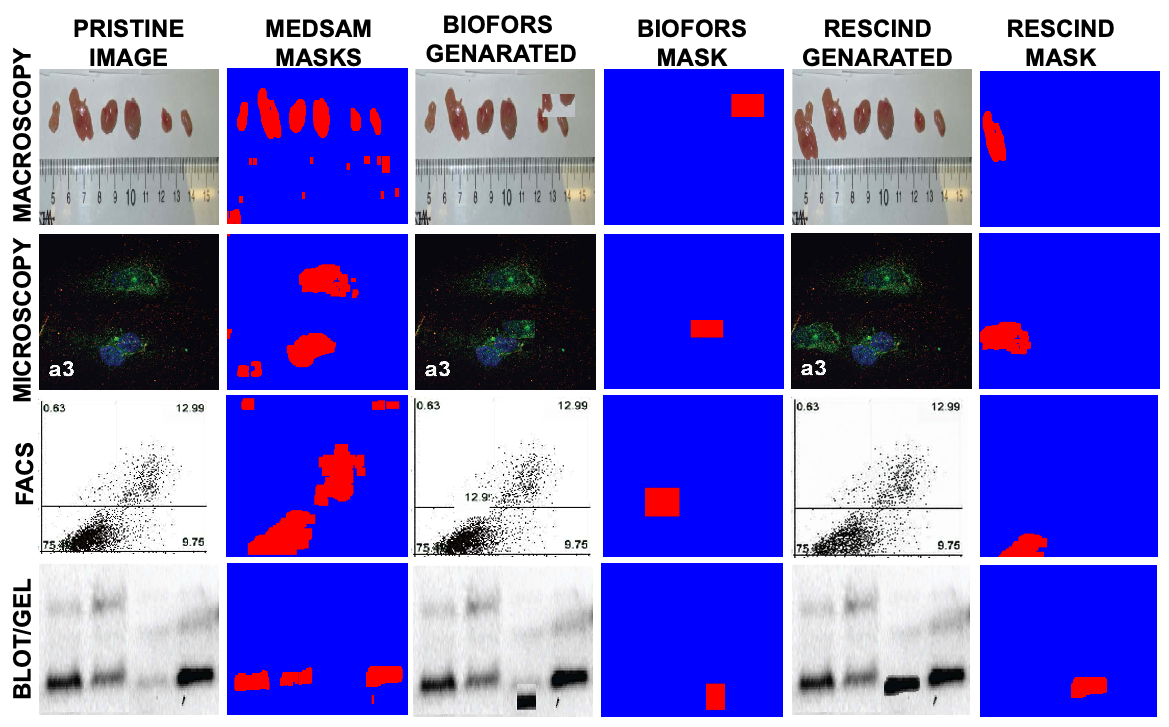}    
    \caption{Forgery synthesis in \textbf{BioFors} vs. \textbf{Rescind} across  biomedical modalities: \textit{macroscopy, microscopy, FACS}, and \textit{blot/gel}. BioFors uses rectangular insertions lacking realism; Rescind leverages MedSAM~\cite{ma2024segment} for irregular mask generation and prompt-guided diffusion for semantically consistent edits. Red = forged; blue = pristine. Eg. Microscopy Prompt: \textit{``Duplicate the fluorescent cell cluster from the highlighted region to simulate manipulation.''}}
    \label{fig:Rescind_intro_fig}
\end{figure}

Scientific image manipulation has emerged as a significant challenge undermining the integrity of biomedical research. Increasing reports of fraudulent practices—including duplication, splicing, and inappropriate alterations—have led to paper retractions, distorted scientific evidence, and erosion of public trust~\cite{bik2016prevalence,stern2014financial}. Compared to natural images, biomedical figures—such as microscopy images, gel electrophoresis bands, and histological slides—exhibit complex textures, modality-specific artifacts, and heterogeneous layouts. These challenges are compounded by the rise of generative AI, which enables sophisticated image fabrication~\cite{gu2022ai,kim2024generative}, making both manual verification and automated detection significantly more difficult.

Recent advances in image forensics have primarily focused on natural images~\cite{wu2019mantra,kwon2021cat,guillaro2023trufor,su2025can}, with limited applicability to scientific figures. Moreover, the scarcity of high-quality, publicly available biomedical forgery datasets limits the development and evaluation of robust AI models for scientific integrity applications. Existing biomedical forensics datasets are often small-scale, lack semantic richness and manipulation diversity, and fail to capture realistic domain-specific artifacts~\cite{sabir2021biofors,moreira2022sila,mandelli2022forensic}, resulting in poor generalization to real-world scientific image forensics.

Despite recent progress in multimodal foundation models, existing vision-language models such as BiomedCLIP~\cite{zhang2024biomedclip} and LLaVA-Med~\cite{li2023llava} are primarily designed for image-text alignment, enabling biomedical image classification, retrieval, and caption generation. However, these models lack the capability to synthesize or manipulate images based on semantic control. Specifically, while they can accurately describe biomedical images and classify figure modalities, they cannot generate new images exhibiting controlled manipulation types such as duplication or splicing. Furthermore, these models are predominantly trained on radiology-focused datasets, making them more effective at interpreting modalities such as radiographs, MRI, and CT scans, but considerably less robust on microscopy, gel electrophoresis (blot), and other molecular biology figures commonly used in biomedical publications. This significant gap limits their applicability in generating realistic forgery datasets or detecting scientific image manipulations within non-radiological biomedical contexts.

To address these challenges, we propose a VLM-guided forgery generation and detection framework for biomedical publications. We introduce \textbf{Rescind}, a large-scale benchmark dataset of synthetically manipulated biomedical images spanning diverse modalities, including microscopy, blot/gel, macroscopy, and flow cytometry (FACS). Rescind is constructed using a VLM-driven semantic control mechanism combined with a diffusion-based generative pipeline to produce realistic forgery types, including duplication, splicing, and region removal. Furthermore, we incorporate a VLM-based automatic verification loop to ensure semantic consistency and quality control of the generated forgeries, addressing a critical gap in existing data curation approaches for scientific image integrity analysis.

In addition to dataset generation, we introduce \textbf{Integscan}, a vision-language architecture for fine-grained biomedical image forgery detection. Integscan integrates structured state-space modeling with lightweight attention mechanisms to encode spatial features, while leveraging vision-language prompts for semantic alignment. Its modular design—including IntegSSM blocks, prompt-conditioned enhancement via IntegBoost, and structured fusion through the Integration module—enables accurate localization of manipulated regions across diverse modalities. To ensure semantic realism in forged samples, we fine-tune diffusion models for region-aware inpainting and incorporate a VLM-based verification loop for filtering. Extensive experiments on Rescind and existing biomedical benchmarks demonstrate that Integscan achieves state-of-the-art performance in both pixel and image-level forgery detection and localization. Our key contributions are:

\begin{itemize}
\item We introduce \textbf{Rescind}, a 600K-scale benchmark for biomedical image forgery detection, comprising three synthetic variants—\textbf{Rescind-I} (image-processed), \textbf{Rescind-G} (generative), and \textbf{Rescind-V} (VLM-guided)—spanning four manipulation types across multiple biomedical modalities.

\item We present the first VLM-guided forgery synthesis pipeline that combines vision-language prompting, generative diffusion, and a post-hoc verification loop using LLaVA-Med to produce semantically aligned and visually realistic biomedical forgeries.


\item We propose \textbf{Integscan}, a state-space model with attention-enhanced visual encoding and prompt-conditioned semantic fusion, achieving state-of-the-art localization across all manipulation tasks and modalities.

\end{itemize}

\section{Related Work: Biomedical Misconduct}
\label{sec:related_work}

\textbf{Limitations of Existing Datasets.} Image manipulation is a growing concern in biomedical publishing, with up to 3.8\% of papers estimated to contain tampered images~\cite{bik2016prevalence}. While synthetic datasets like BioFors~\cite{sabir2021biofors}, Mandelli et al.~\cite{mandelli2022forensic}, and M3D-Synth~\cite{zingarini2024m3dsynth} have supported detection efforts, they are constrained by modality (e.g., Western blots, CT scans), lack pixel-level annotations, or simulate manipulations using simplistic heuristics. To overcome these limitations, we introduce \textbf{Rescind}, a general-purpose biomedical forgery dataset with fine-grained masks spanning microscopy, blot/gel, macroscopy, and FACS. Rescind includes semantically meaningful, irregular manipulations grounded in real figures, making it well-suited for training and evaluation.

\textbf{Towards General-Purpose Forgery Detection.} Despite advances in biomedical segmentation~\cite{shamshad2023transformers, heidari2023hiformer, wu2022fat} and diagnosis~\cite{esteva2021deep}, biomedical forensics remains underexplored due to limited data and task diversity. Early duplication detection methods~\cite{koppers2017towards, bucci2018automatic, acuna2018bioscience} based on hand-crafted features perform poorly in domains with subtle textures and structural repetition. Deep models~\cite{wu2019mantra, guillaro2023trufor, su2025can} have improved performance but often rely on splicing assumptions and fail under complex internal forgeries. While BioFors~\cite{sabir2021biofors} introduced EDD, IDD, and CSTD benchmarks, most models—e.g., MONet~\cite{sabir2022monet}—focus on EDD alone. To this end, we propose \textbf{IntegScan}, a vision-language-guided state-space model that jointly localizes source and manipulated regions across all three BioFors tasks, offering improved generalization to diverse biomedical forgery types.

\begin{figure*}[t]
    \centering
    \includegraphics[width=0.9\linewidth]{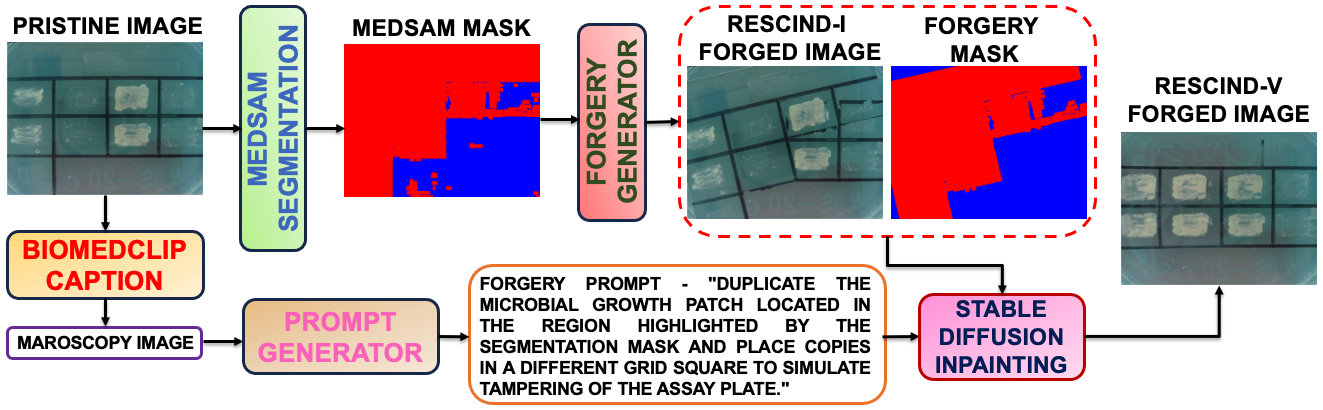}
    \caption{\textbf{Rescind-V Curation Pipeline}. Given a \textit{pristine biomedical image}, BiomedCLIP~\cite{zhang2024biomedclip} generates a modality-aware caption (e.g., \textit{macroscopy image}). MedSAM~\cite{ma2024segment} produces a binary segmentation mask that identifies regions of interest. A \textbf{Forgery Generator} creates a Rescind-I forged image and corresponding forgery mask by duplicating or manipulating content within the masked region. A \textbf{Prompt Generator} formulates a semantic forgery prompt based on the predicted modality, explicitly referencing the MedSAM-highlighted region. Finally, a Stable Diffusion inpainting model~\cite{Rombach_2022_CVPR} synthesizes the \textbf{Rescind-V forged image} using the \textit{forgery mask} and \textit{forgery prompt} as conditional inputs.}
    \label{fig:Rescind_V_forgery_pipeline}
\end{figure*}

\textbf{VLMs for Biomedical Images.}
VLMs like CLIP~\cite{radford2021learning} and its biomedical variants—BiomedCLIP~\cite{zhang2024biomedclip}, OpenBioMed~\cite{OpenBioMed_code}, and LLaVA-Med~\cite{li2023llava}—excel in classification, retrieval, and captioning. However, they are not designed to detect manipulation or exposed to forged data. We fine-tune VLMs on Rescind with manipulation-aware prompts, enabling semantically conditioned forgery detection and enhancing both interpretability and realism.

\textbf{State Space Models.}
Structured State Space Models (SSMs), such as S4~\cite{gu2021efficiently}, Mamba~\cite{gu2023mamba}, and Mamba-2~\cite{mamba2}, model long-range dependencies efficiently and are increasingly viewed as generalizations of Transformers~\cite{han2024demystify}. Vision extensions like ViMamba~\cite{liu2024vmamba} have shown promise in segmentation, but remain generic. We introduce the first SSM-based model for biomedical image forensics, leveraging structured recurrence and attention to achieve high accuracy on complex forgery types.

\section{Rescind Dataset Construction}
\label{sec:rescind}

We introduce Rescind, a large-scale dataset of semantically controlled and visually realistic forgeries in biomedical images. Rescind is curated through a multi-stage pipeline that integrates vision-language understanding, region segmentation, and guided image generation to produce high-fidelity forgeries while maintaining biological plausibility. Figure~\ref{fig:Rescind_V_forgery_pipeline} provides an overview of the complete generation pipeline.

\subsection{Pristine Image Collection}

We begin by curating a diverse set of authentic biomedical figures from publicly available and ethically approved datasets. Specifically, 1,031 Portable Document Format (PDF) files containing raw annotations of suspicious scientific figures were obtained from~\cite{bik2016prevalence}, of which 335 contained at least one manipulated figure. Following the approach in~\cite{sabir2021biofors}, we extracted 17,269 manipulated images (\textbf{Rescind-T}) and 30,536 unmanipulated pristine images (\textbf{Rescind-P}) from these figures. The images span a range of biomedical modalities and are categorized into four major classes based on semantics/visual appearance as illustrated in first column of Figure~\ref{fig:Rescind_intro_fig}: 

\textbf{Microscopy} comprises images captured using light or electron microscopes, typically depicting stained cells and tissues. These images vary widely in color and structure due to differences in biological source and staining techniques, contributing to their visual heterogeneity.

\textbf{Blots/Gels} include western, northern, and southern blot images and electrophoresis gels used for analyzing proteins, RNA, and DNA. Despite targeting different biomolecules, they exhibit similar visual formats such as bands and lanes, often indistinguishable without textual labels.

\textbf{Flow Cytometry (FACS)} images resemble scatter plots, showing cell or particle distributions captured via fluorescence-based experiments. Though visually simple, they encode biologically meaningful cell population data.

\textbf{Macroscopy} includes biomedical images captured without magnification, such as gross anatomical views, surgical fields, or laboratory setups, and exhibits high variability in structure and context.

These modality-aware, richly annotated pristine images form the foundation for generating both authentic and synthetic samples in the Rescind dataset.

\subsection{Synthetic Forgery Generation}

Building on the modality-aware pristine figures in \textbf{Rescind-P}, we construct three categories of synthetically manipulated images, each simulating different manipulation strategies commonly encountered in biomedical research. These subsets—\textbf{Rescind-I}, \textbf{Rescind-G}, and \textbf{Rescind-V}—enable diverse and controllable image forgery types ranging from low-level transformations to semantically guided edits.

\noindent\textbf{Rescind-I (Image-processed forgeries)} contains forged images created by applying image-level transformations to segmented regions extracted from Rescind-P using MedSAM~\cite{ma2024segment}. These transformations include scaling, rotation, shifting, and duplication of image patches, which are either repositioned within the same image (Internal Duplication Detection - IDD) or blended into new host images using MGMatting~\cite{yu2021mask} (External Duplication Detection - EDD). Unlike prior datasets with rectangular masks (e.g., BioFors~\cite{sabir2021biofors}), Rescind-I employs irregular masks that better reflect natural biological structures, improving realism and localization difficulty.

\noindent\textbf{Rescind-G (Generative forgeries)} builds upon Rescind-P and Rescind-I by introducing forgery types generated via image-to-image generative models. Specifically, we apply Stable Diffusion inpainting~\cite{Rombach_2022_CVPR}, Pix2Pix GAN~\cite{isola2017image}, and CycleGAN~\cite{zhu2017unpaired} to MedSAM-derived masks. These methods generate realistic content in place of or in addition to existing regions, simulating removal or substitution forgeries. The generated patches are applied either to the original region defined by MedSAM or to new locations introduced in Rescind-I, allowing spatial and contextual variability.

\noindent\textbf{Rescind-V (VLM-guided semantic forgeries)} introduces semantic control into forgery synthesis by leveraging vision-language models (VLMs). Each image in Rescind-P is first captioned using BiomedCLIP~\cite{zhang2024biomedclip} to extract modality-aware descriptions. MedSAM~\cite{ma2024segment} is then applied to segment regions of interest, which are paired with curated natural language prompts describing plausible manipulations (e.g., \textit{``duplicate the protein band gel lane identified by the mask and replicate it in a new position to fabricate consistent results across experiments''} or \textit{``Erase the mitotic cell cluster from the highlighted area to simulate data suppression in the microscopy image.''}).

\begin{table}[t]
\centering
\small 
\setlength{\tabcolsep}{1.3pt}    
\begin{tabular}{l|c|c|c|c|c}
\hline
\textbf{Subset} & \textbf{Blot} & \textbf{Micro} & \textbf{Macro} & \textbf{FACS} & \textbf{Total} \\
\hline
\textbf{Rescind-P (Train)}      & 19,105 & 10,458 & 555  & 418  & 30,536 \\
\textbf{Rescind-T (Test)}       & 8,335  & 7,652  & 639  & 643  & 17,269 \\
\hline
\textbf{Rescind-I (EDD)}        & 40,002 & 25,002 & 2,502 & 2,502 & 70,008 \\
\textbf{Rescind-I (IDD)}        & 9,999  & 6,501  & 1,251 & 1,251 & 19,002 \\
\textbf{Rescind-I (CSTD)}       & 9,999  & 6,501  & 1,251 & 1,251 & 19,002 \\
\textbf{Rescind-I (Removal)}    & 40,002 & 25,002 & 2,502 & 2,502 & 70,008 \\
\hline
\textbf{Rescind-G (EDD)}        & 40,002 & 25,002 & 2,502 & 2,502 & 70,008 \\
\textbf{Rescind-G (IDD)}        & 9,999  & 6,501  & 1,251 & 1,251 & 19,002 \\
\textbf{Rescind-G (CSTD)}       & 9,999  & 6,501  & 1,251 & 1,251 & 19,002 \\
\textbf{Rescind-G (Removal)}    & 40,002 & 25,002 & 2,502 & 2,502 & 70,008 \\
\hline
\textbf{Rescind-V (EDD)}        & 40,002 & 25,002 & 2,502 & 2,502 & 70,008 \\
\textbf{Rescind-V (IDD)}        & 9,999  & 6,501  & 1,251 & 1,251 & 19,002 \\
\textbf{Rescind-V (CSTD)}       & 9,999  & 6,501  & 1,251 & 1,251 & 19,002 \\
\textbf{Rescind-V (Removal)}    & 40,002 & 25,002 & 2,502 & 2,502 & 70,008 \\
\hline
\textbf{Rescind-Total}          & 327,446 & 207,128 & 23,712 & 23,579 & \textbf{581,865} \\
\hline
\end{tabular}
\caption{Rescind Summary: pristine(P), real test(T), and synthetic subsets—Image-processed(I), Generative(G), VLM-guided(V); Task types: EDD (external duplication), IDD (internal duplication), CSTD (cut/sharp transition), Removal.}
\label{tab:rescind_full_summary}
\end{table}

These prompts are generated by a Prompt Generator, which conditions on the BiomedCLIP-derived caption and incorporates manipulation patterns inspired by the forensic taxonomy of notes outlined in~\cite{bik2016prevalence} and real-world examples from retracted publications. To guide prompt construction, we also studied the notes and annotations accompanying detected forgeries in retracted papers, enabling the generation of realistic semantic prompts that reflect tampering intent from a perpetrator's perspective.

A Forgery Generator then uses the MedSAM-produced mask to create an initial Rescind-I forged image and corresponding forgery mask. While Rescind-I introduces structural changes, these forgeries are not always semantically coherent. To overcome this, Rescind-V employs Stable Diffusion inpainting~\cite{Rombach_2022_CVPR}, conditioned on the pristine image, the forgery mask (which guides the inpainting region), and the semantic prompt, to synthesize realistic and contextually consistent forgeries. This subset emphasizes language-aligned, modality-consistent manipulation and supports the development and evaluation of VLM-sensitive forgery detection models.

Together, Rescind-I, Rescind-G, and Rescind-V form a unified benchmark for image integrity, spanning low-level duplication, generative manipulation, and semantically guided forgery synthesis, as summarized in Table~\ref{tab:rescind_full_summary}.

\section{Proposed Method}
\label{sec:method}

Integscan is a VLM architecture for fine-grained biomedical image forgery detection. It integrates structured spatial encoding and prompt-conditioned semantic alignment to localize manipulated regions across diverse modalities.

\subsection{Integscan Pipeline Overview}

\begin{figure}[t]
    \centering
    \includegraphics[width=1.0\linewidth]{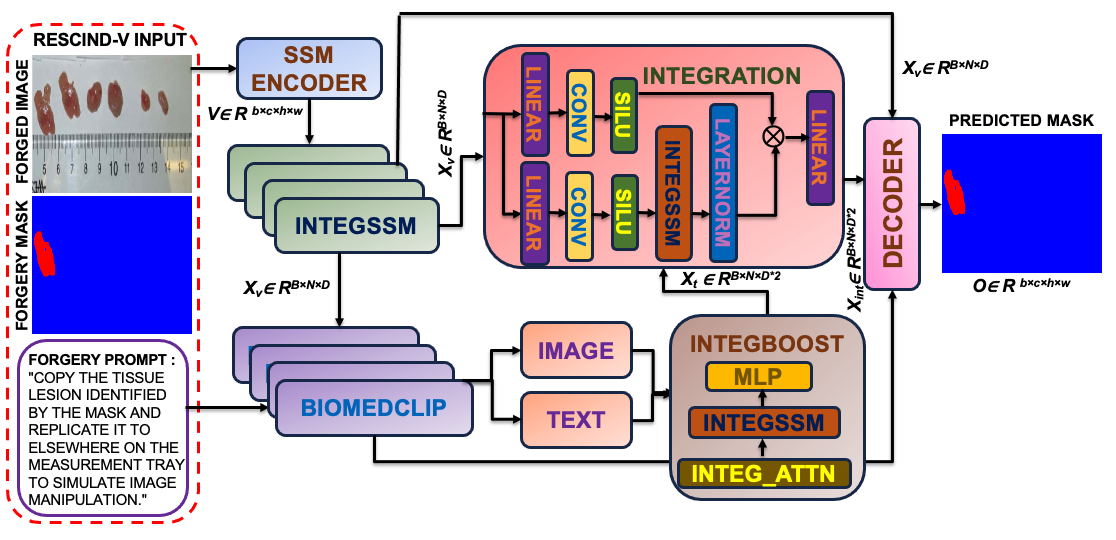}
    \caption{Integscan Model Architecture Overview}
    \label{fig:IntegScan_fig}
\end{figure}

As illustrated in Figure~\ref{fig:IntegScan_fig}, given a forged Rescind-V image $V \in \mathbb{R}^{b \times c \times h \times w}$, its corresponding forgery mask, and an associated prompt, \textbf{Integscan} predicts a binary manipulation mask $\hat{O} \in \mathbb{R}^{b \times c \times h \times w}$, where $b$, $c$, $h$, and $w$ denote batch size, channels, height, and width, respectively. A structured state-space encoder first extracts visual features from $V$, which are processed by stacked \textbf{IntegSSM} blocks to generate flattened spatial tokens $X_v \in \mathbb{R}^{b \times N \times D}$, where $N = H_f \times W_f$ is the number of downsampled spatial tokens and $D$ is the embedding dimension.

To incorporate modality-specific semantics, we leverage BiomedCLIP~\cite{zhang2024biomedclip} to retrieve relevant prompts and embed them alongside visual features. These embeddings are passed to the \textbf{IntegBoost} module, which performs semantic enhancement using structured attention and state-space reasoning to produce enriched prompt-aware features $X_t \in \mathbb{R}^{b \times N \times 2D}$. The enhanced text features are fused with the visual tokens using the \textbf{Integration} module, yielding a unified representation $X_{\text{int}} \in \mathbb{R}^{b \times N \times 2D}$. Finally, $X_{\text{int}}$ is reshaped and passed through a lightweight convolutional decoder comprising a \texttt{ReLU}-activated $3{\times}3$ layer followed by a $1{\times}1$ projection to generate the dense manipulation mask. We detail the core visual encoding and attention components of Integscan below:

\subsection{IntegSSM Block: Structured Visual Encoding}
\label{sec:integssm_block}

\begin{figure}[t]
    \centering
    \includegraphics[width=0.6\linewidth]{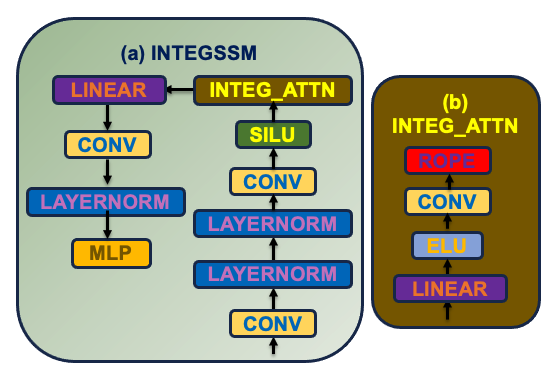}
    \caption{(a) IntegSSM with state-space modeling; (b) INTEG\_ATTN with rotary encoding, channel-aware attention.}
    \label{fig:IntegSSM_fig}
\end{figure}

The \textbf{IntegSSM} block forms the backbone of the visual encoder in Integscan. It sequentially integrates convolutional and attention-based operations to encode spatial structure at multiple levels of granularity. As shown in Figure~\ref{fig:IntegSSM_fig}(a), the input tokens first pass through a \texttt{CONV} layer to extract local spatial features. This is followed by two successive \texttt{LAYERNORM} layers, interleaved with additional \texttt{CONV} operations to refine the representation. A \texttt{SILU} activation introduces non-linearity before the tokens are passed into the \texttt{INTEG\_ATTN} submodule (detailed in Section~\ref{sec:integ_attn}), which models long-range spatial dependencies using linear attention with rotary positional encoding. The output is then projected via a \texttt{LINEAR} layer and passed through another \texttt{CONV} $\rightarrow$ \texttt{LAYERNORM} $\rightarrow$ \texttt{MLP} sequence to further mix token interactions across space and channels. The final output is produced by a concluding \texttt{LINEAR} projection. This block design enables efficient extraction of hierarchical features while preserving both local and global contextual cues.

\subsection{Integ\_Attn: Structured Linear Attention via SSM}
\label{sec:integ_attn}

At the heart of the IntegSSM block lies the \texttt{Integ\_Attn} submodule, which implements structured linear attention based on discretized State Space Models (SSMs). This formulation allows efficient and scalable modeling of long-range spatial dependencies through learned recurrence and positional priors. The continuous-time SSM is defined as:
\begin{equation}
h'(t) = \mathbf{A}h(t) + \mathbf{B}x(t), \quad y(t) = \mathbf{C}h(t)
\label{eq:ssm-continuous}
\end{equation}
To integrate this into deep networks, we apply zero-order hold (ZOH) discretization~\cite{gu2021efficiently}, resulting in:
\begin{equation}
h_k = \bar{\mathbf{A}} h_{k-1} + \bar{\mathbf{B}} x_k, \quad y_k = \bar{\mathbf{C}} h_k + \bar{\mathbf{D}} x_k
\label{eq:ssm-discrete}
\end{equation}
where $\bar{\mathbf{A}}, \bar{\mathbf{B}}, \bar{\mathbf{C}}$ are learned discrete parameters and $\bar{\mathbf{D}}$ provides a residual shortcut.

In practice, \texttt{Integ\_Attn} approximates these dynamics through a series of differentiable steps. The input is projected into query and key vectors, activated with ELU and offset by a bias term to ensure positive gating. Rotary positional encoding (RoPE) is applied to both $q$ and $k$ to encode spatial structure in a recurrent form. Token-wise recurrence is simulated by computing attention weights through interactions between $q$ and a compressed average of $k$. These weights are used to aggregate value tokens, which are further enhanced with a depthwise convolutional layer—analogous to the residual $\bar{\mathbf{D}}x_k$ term in Equation~\ref{eq:ssm-discrete}. By combining structured recurrence, positional priors, and local spatial enhancement, \texttt{Integ\_Attn} provides a lightweight yet expressive mechanism for structured attention within Integscan.

\subsection{IntegBoost: Prompt Enhance via Visual Context}
\label{sec:integboost}

The \textbf{IntegBoost} module, shown in Figure~\ref{fig:IntegScan_fig}, enhances prompt embeddings by conditioning them on visual priors extracted from the forged image. Unlike previous methods that treat text as static queries, IntegBoost dynamically refines the prompts by modeling their interaction with image features. It consists of a lightweight attention module, a structured IntegSSM block, and an MLP, each equipped with residual connections. The attention layer first captures contextual alignment between the prompt and visual tokens, while the IntegSSM block further integrates spatial priors into the prompt representation. This hybrid design improves semantic grounding, allowing the prompt to effectively guide manipulation localization across diverse biomedical modalities.

\subsection{Integration Module: Visual–Prompt Feat. Fusion }
\label{sec:integration}

The \textbf{Integration} module, shown in Figure~\ref{fig:IntegScan_fig}, fuses visual features from IntegSSM with prompt-enhanced representations from IntegBoost to produce a unified token sequence for manipulation localization. Rather than relying on cross-attention over concatenated tokens—which scales poorly with sequence length—the module adopts a structured design based on linear projections, convolutional refinement, and state-space modeling. Both inputs are first projected into a shared embedding space and reshaped into spatial grids. Each stream is individually processed by \texttt{SiLU}-activated convolution layers to capture local structure, then element-wise summed to encourage early feature interaction. The result is passed through an \texttt{IntegSSM} block, which models spatial dependencies across the fused representation via structured recurrence. A final normalization and linear projection produce the fused embedding $X_{\text{int}} \in \mathbb{R}^{b \times N \times 2D}$ for downstream decoding. Inspired by adapter-based learning~\cite{houlsby2019parameter}, this approach allows visual and prompt features to interact meaningfully without fully fine-tuning large pretrained encoders. The state-space fusion captures both long-range token relationships and local visual cues while maintaining computational efficiency. This makes Integration a lightweight yet expressive module for prompt-guided segmentation.

\section{Experimental Evaluation}
\label{sec:exp_eval}

\subsection{Rescind Quality Assessment \& VLM Validation}

To ensure the realism and utility of the Rescind dataset, we conduct both perceptual and semantic quality evaluations of the generated forgeries.

\noindent\textbf{Visual Quality Evaluation.} We assess image fidelity using both no-reference and full-reference image quality metrics. For forged images alone, we employ no-reference metrics such as NIQE~\cite{ying2020patches} and PIQE~\cite{wang2023exploring} to evaluate perceptual naturalness in the absence of ground truth. For comparing forged images against their pristine counterparts, we use full-reference metrics including SSIM~\cite{wang2004image} and LPIPS~\cite{zhang2018unreasonable} to measure structural similarity and perceptual distance. To further eliminate trivially unrealistic samples, we train a binary image classifier to detect synthetic forgeries, filtering out generated samples with high fake-confidence scores to maintain dataset integrity. Table~\ref{tab:rescind_iqa} reports the results of these quality assessments on the synthetic subsets of the Rescind dataset. Among all subsets, \textbf{Rescind-V} achieves the best scores across all metrics—lowest NIQE and PIQE, highest SSIM, and lowest LPIPS—indicating that vision-language model (VLM)-guided forgeries are the most visually plausible and semantically coherent. This underscores the importance of incorporating modality-aware semantic control in generating high-fidelity biomedical forgeries.

\begin{table}[t]
    \centering
    \small
    \setlength{\tabcolsep}{4pt}
    \begin{tabular}{l|c|c|c|c}
        \toprule
        \textbf{Metric} & \textbf{Type} & \textbf{Rescind-I} & \textbf{Rescind-G} & \textbf{Rescind-V} \\
        \midrule
        NIQE ($\downarrow$) & No-Ref & 4.23 & 3.79 & \textbf{3.31} \\
        PIQE ($\downarrow$) & No-Ref & 38.7 & 30.5 & \textbf{25.9} \\
        \midrule
        SSIM ($\uparrow$) & Full-Ref & 0.827 & 0.871 & \textbf{0.894} \\
        LPIPS ($\downarrow$) & Full-Ref & 0.204 & 0.143 & \textbf{0.116} \\
        \bottomrule
    \end{tabular}
    \caption{Image Quality Assessment (IQA) on Rescind subsets. 
    $\uparrow$ indicates higher is better; $\downarrow$ indicates lower is better.}
    \label{tab:rescind_iqa}
\end{table}

\begin{figure}[t]
    \centering
    \includegraphics[width=0.85\linewidth]{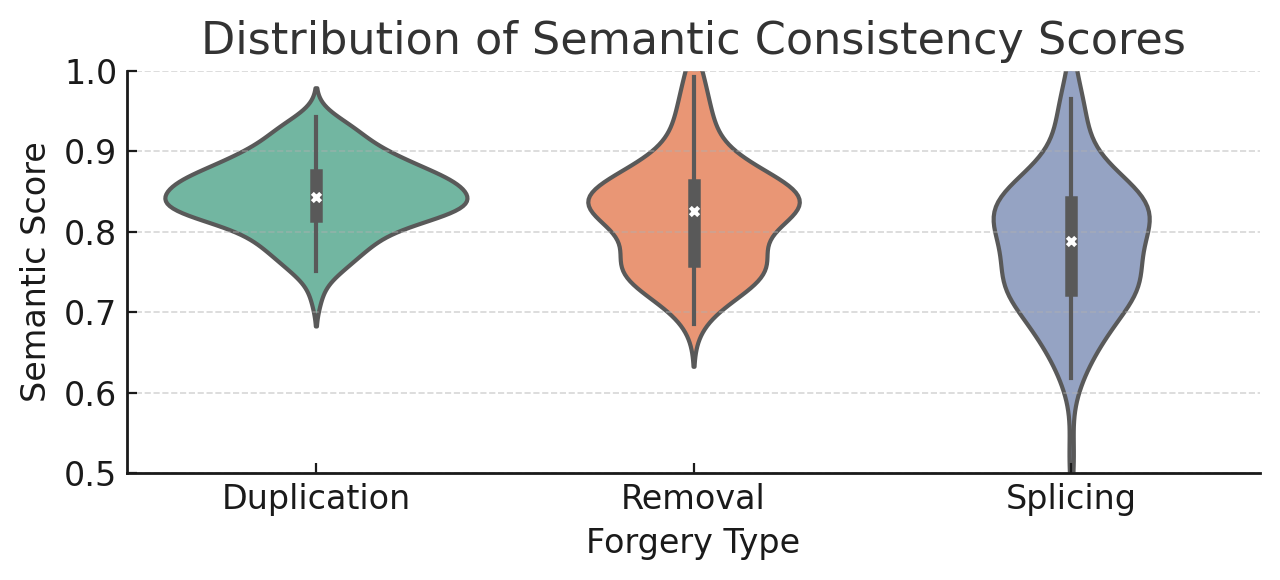}
    \caption{Semantic Consistency Score across forgery types, using LLaVA-Med on Rescind-I, Rescind-G, Rescind-V.}
    \label{fig:semantic_violin}
\end{figure}

\noindent\textbf{Semantic Consistency Verification.} Beyond visual realism, we ensure that each manipulation aligns with its intended semantic objective. To this end, we employ LLaVA-Med~\cite{li2023llava} in a post-hoc verification loop. Each generated image is captioned by the VLM using a modality-specific manipulation prompt (e.g., ``image with duplicated blot band''), and the output is compared against the expected semantic intent. Images falling below a predefined alignment threshold are discarded. The remaining forgeries are annotated with a \textit{Semantic Consistency Score}, which quantifies alignment quality for downstream analysis. Figure~\ref{fig:semantic_violin} shows score distributions across forgery types, with duplication exhibiting the highest semantic fidelity.

\subsection{Integscan Results and Analysis}
\label{sec:results}

Integscan achieves state-of-the-art performance on the Rescind-T test set, containing real-world duplication forgeries from retracted biomedical publications. As shown in Tables~\ref{tab:idd_only}-\ref{tab:edd_only}, it consistently outperforms prior methods across all biomedical modalities and tasks, including both External and Internal Duplication Detection (EDD and IDD). Unlike MONet~\cite{sabir2022monet}, which is limited to EDD, Integscan delivers robust performance across all categories.

Traditional keypoint-based detectors (e.g., SIFT~\cite{lowe2004distinctive}, ORB~\cite{rublee2011orb}) and handcrafted feature methods (e.g., DCT~\cite{fridrich2003detection}, Zernike~\cite{ryu2010detection}) perform poorly on biomedical images due to structural repetition. While dense matching methods like DF-ZM and DF-PCT~\cite{cozzolino2015efficient} outperform earlier techniques, their reliance on low-level features limits their generalization. Deep models such as ManTra-Net~\cite{wu2019mantra}, Trufor~\cite{guillaro2023trufor}, and MONet struggle to handle duplication-based tampering, particularly when confined to target-only prediction.


Integscan uniquely localizes both source and target regions in duplication forgeries, including challenging internal duplication scenarios. While competing models often misclassify biomedical textures or detect only superficial manipulation artifacts, Integscan accurately identifies contextually aligned forgeries with high spatial precision. This is achieved by combining structured attention with state-space modeling and prompt-guided semantics, enabling robust visual reasoning across diverse biomedical modalities.

\subsection{Ablation Studies}

\begin{table}[t]
\centering
\footnotesize
\small
\setlength{\tabcolsep}{1.5pt}
\begin{tabular}{l|cc|cc|cc|cc}
\toprule
& \multicolumn{8}{c}{\textbf{Internal Duplication Detection (IDD)}} \\
\cmidrule(lr){2-9}
\textbf{Method} & \multicolumn{2}{c|}{Microscopy} & \multicolumn{2}{c|}{Blot/Gel} & \multicolumn{2}{c|}{Macroscopy} & \multicolumn{2}{c}{Combined} \\
& Image & Pixel & Image & Pixel & Image & Pixel & Image & Pixel \\
\midrule
DF-ZM & \underline{0.764} & 0.197 & 0.515 & 0.449 & 0.573 & 0.478 & 0.564 & 0.353 \\
DF-PCT & \underline{0.764} & \underline{0.202} & \underline{0.503} & \underline{0.466} & \underline{0.712} & \underline{0.487} & \underline{0.569} & \underline{0.364} \\
DF-FMT & 0.638 & 0.167 & 0.480 & 0.400 & 0.495 & 0.458 & 0.509 & 0.316 \\
BusterNet & 0.183 & 0.178 & 0.226 & 0.076 & 0.021 & 0.106 & 0.269 & 0.107 \\
ManTra & 0.316 & 0.194 & 0.317 & 0.094 & 0.272 & 0.262 & 0.335 & 0.183 \\
TruFor & 0.321 & 0.197 & 0.324 & 0.099 & 0.245 & 0.252 & 0.336 & 0.197 \\
\midrule
\textbf{Integscan} & \textbf{0.906} & \textbf{0.604} & \textbf{0.768} & \textbf{0.694} & \textbf{0.925} & \textbf{0.764} & \textbf{0.793} & \textbf{0.614} \\
\bottomrule
\end{tabular}
\caption{MCC scores for IDD on Rescind-T test-set for both Image and Pixel levels; best \textbf{bold}, second-best \underline{underlined}}
\label{tab:idd_only}
\end{table}

\textbf{Semantic Prompt Importance.} We ablate the vision-language component by generating Rescind-V samples without BiomedCLIP-guided prompts. Instead, we apply random or generic prompts (e.g., "remove a region"). Detection models trained on prompt-less samples perform worse on semantically consistent manipulations, confirming the critical role of prompt conditioning for generating realistic, language-aligned forgeries.

\noindent\textbf{Modality-Specific Generalization.} We perform a leave-one-modality-out experiment where models are trained on three modalities and evaluated on the held-out fourth. This reveals cross-modality generalization gaps and highlights the need for modality-aware modeling. For example, models trained without FACS perform poorly on FACS test samples, indicating that visually simple modalities require explicit inclusion in training.

\noindent\textbf{Integscan Ablation.} To understand the contribution of each component in the \textbf{Integscan} architecture, we conduct systematic ablations on the Rescind-T EDD test set (Table~\ref{tab:ablation_integscan}). Removing the \textbf{IntegBoost} module, which enhances text embeddings through image-conditioned attention, leads to a noticeable drop in F1 score and IoU, highlighting the importance of modality-aware prompt refinement. Excluding the \textbf{Integration} fuser, which bridges VLM and VFM features, reduces the model's ability to semantically align predictions, resulting in degraded performance. When the \textbf{IntegSSM} visual backbone is replaced with a standard convolutional encoder, the model's spatial reasoning capacity is significantly reduced, confirming the advantage of long-range attention via structured state-space modeling. Finally, bypassing the prompt-conditioning process and using a generic prompt across all images leads to a marked decline in localization performance, verifying the necessity of semantic prompts tailored to the image modality. These results collectively validate the design choices of Integscan in achieving precise and robust forgery localization.

\begin{table}[t]
\centering
\small
\setlength{\tabcolsep}{0.3pt}
\renewcommand{\arraystretch}{1.5}
\begin{tabular}{l|c|c|c|c|c|c|c|c|c|c}
\toprule
& \multicolumn{10}{c}{\textbf{External Duplication Detection (EDD)}} \\
\cmidrule(lr){2-11}
\textbf{Method} & \multicolumn{2}{c}{\textbf{Micro}} & 
\multicolumn{2}{c}{\textbf{Blot/Gel}} & 
\multicolumn{2}{c}{\textbf{Macro}} & 
\multicolumn{2}{c}{\textbf{FACS}} & 
\multicolumn{2}{c}{\textbf{Comb.}} \\
& Img & Px & Img & Px & Img & Px & Img & Px & Img & Px \\
\midrule
SIFT & 0.180 & 0.146 & 0.113 & 0.148 & 0.130 & 0.194 & 0.110 & 0.073 & 0.142 & 0.132 \\
ORB  & 0.319 & 0.342 & 0.087 & 0.127 & 0.126 & 0.226 & 0.269 & 0.187 & 0.207 & 0.252 \\
BRIEF & 0.275 & 0.277 & 0.058 & 0.102 & 0.135 & 0.169 & 0.244 & 0.188 & 0.180 & 0.202 \\
DF-ZM & 0.422 & 0.425 & 0.161 & 0.192 & \underline{0.285} & \underline{0.256} & \underline{0.540} & \underline{0.504} & 0.278 & 0.324 \\
DMVN & 0.342 & 0.242 & 0.430 & 0.261 & 0.238 & 0.185 & 0.282 & 0.164 & 0.310 & 0.244 \\
ManTra & 0.347 & 0.244 & 0.449 & 0.287 & 0.275 & 0.202 & 0.337 & 0.186 & 0.351 & 0.231 \\
TruFor & 0.356 & 0.261 & 0.468 & 0.299 & 0.214 & 0.357 & 0.349 & 0.189 & 0.409 & 0.317 \\
MONet & \underline{0.435} & \underline{0.398} & \underline{0.520} & \underline{0.507} & 0.262 & 0.221 & 0.356 & 0.313 & \underline{0.438} & \underline{0.410} \\
\midrule
\textbf{Ours} & \textbf{0.821} & \textbf{0.556} & \textbf{0.755} & \textbf{0.662} & \textbf{0.818} & \textbf{0.654} & \textbf{0.726} & \textbf{0.507} & \textbf{0.776} & \textbf{0.604} \\
\bottomrule
\end{tabular}

\caption{MCC scores for EDD on Rescind-T test-set for both Image and Pixel levels; best \textbf{bold}, second-best \underline{underlined}}
\label{tab:edd_only}
\end{table}

\begin{table}[t]
\centering
\small
\setlength{\tabcolsep}{2.5pt}
\renewcommand{\arraystretch}{1.1}
\begin{tabular}{l|c|c|c|c}
    \toprule
    \textbf{Training Configuration} & \textbf{Precision} & \textbf{Recall} & \textbf{F1} & \textbf{IoU} \\
    \midrule
    w/o BiomedCLIP Prompts   & 75.92 & 74.18 & 75.04 & 61.20 \\
    w/o Microscopy (test Mic.) & 72.35 & 68.49 & 70.36 & 55.81 \\
    w/o Blots/Gels (test Blot/Gel) & 73.12 & 71.18 & 72.14 & 58.26 \\
    w/o FACS (test FACS)      & 66.88 & 60.57 & 63.40 & 49.75 \\
    w/o Macroscopy (test Macro) & 70.15 & 67.03 & 68.55 & 54.39 \\
    \midrule
    \textbf{Full Dataset}   & \textbf{79.82} & \textbf{78.05} & \textbf{78.92} & \textbf{63.48} \\
    \bottomrule
\end{tabular}
\caption{Dataset ablation on Rescind-V using Integscan.}
\label{tab:ablation_rescindv_dataset}
\end{table}

\begin{table}[t]
\centering
\small
\setlength{\tabcolsep}{2.5pt}
\renewcommand{\arraystretch}{1.1}
\begin{tabular}{p{3.4cm}|c|c|c|c}
    \hline
    \textbf{Model Variant} & \textbf{Precision} & \textbf{Recall} & \textbf{F1 Score} & \textbf{IoU} \\
    \hline
    w/o IntegBoost              & 94.62 & 92.88 & 93.74 & 88.86 \\
    w/o Integration Module      & 92.19 & 90.72 & 91.45 & 85.33 \\
    w/o Prompt Conditioning     & 91.74 & 89.31 & 90.51 & 83.94 \\
    IntegSSM → CNN  & 90.28 & 88.75 & 89.51 & 81.80 \\
    \hline
    \textbf{Integscan (Full)}   & \textbf{97.83} & \textbf{96.94} & \textbf{97.38} & \textbf{94.96} \\
    \hline
\end{tabular}
\caption{Ablation study on the Rescind-T EDD images.}
\label{tab:ablation_integscan}
\end{table}

\subsection{Robustness Analysis}

\begin{figure}[t]
    \centering
    \includegraphics[width=1.0\linewidth]{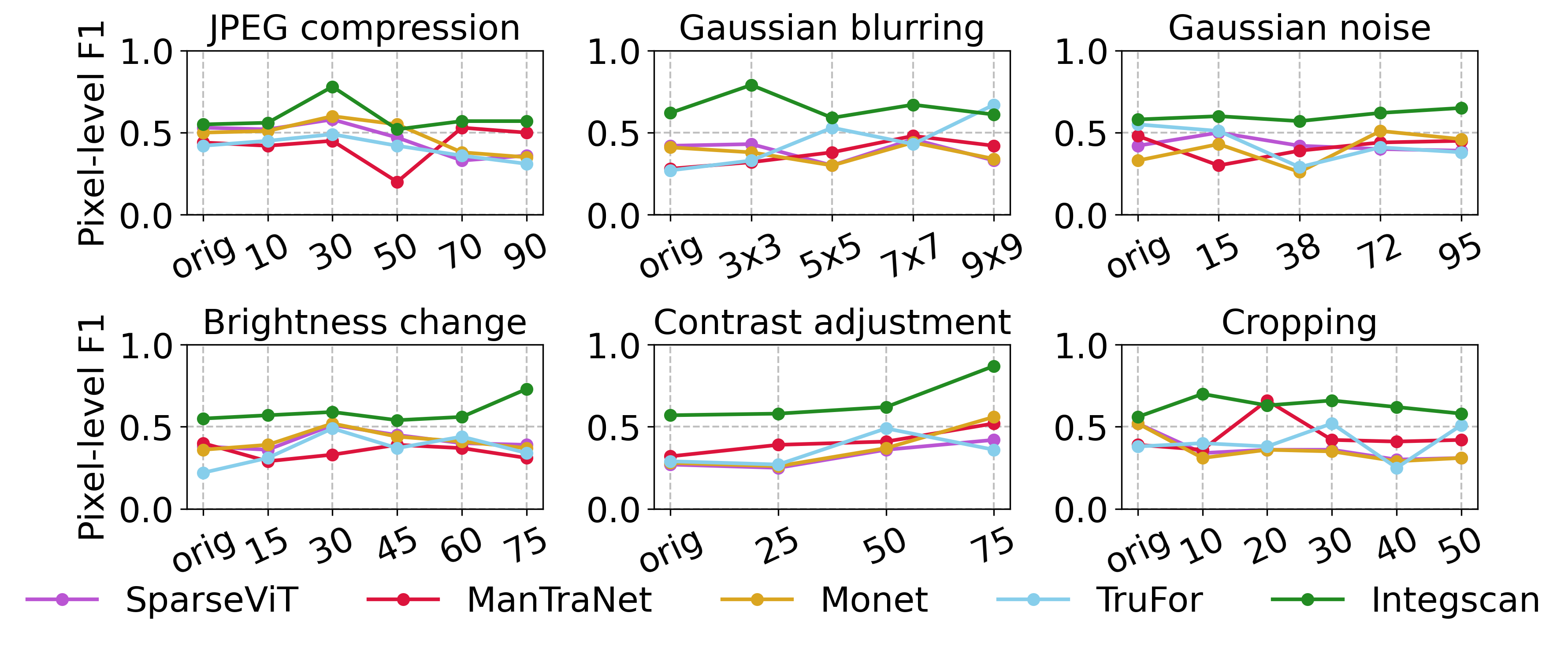}
    \caption{Robustness under attacks on Rescind-V}
    \label{fig:robustness_fig}
\end{figure}

To assess the robustness of forgery detection models under real-world degradations, we evaluate all baselines on the \textbf{Recind-V} dataset across six common perturbations—\textit{JPEG compression}, \textit{Gaussian blurring}, \textit{Gaussian noise}, \textit{brightness shift}, \textit{contrast adjustment}, and \textit{cropping}—following the protocol in~\cite{wu2018busternet, wu2019mantra}. Figure~\ref{fig:robustness_fig} presents the pixel-level F1 scores for five representative detection methods under increasing perturbation severity. \textbf{Integscan} demonstrates consistently superior performance across all attack types, exhibiting graceful degradation even under heavy distortions. In contrast, existing methods such as SparseViT, MVSS, and TruFor show significant performance drops under geometric and low-level corruptions. These results confirm that Integscan is not only accurate but also highly robust to post-manipulation operations commonly found in scientific image tampering.

\section{Limitations and Future Work}
\label{sec:limitations}

While Integscan achieves strong performance across diverse biomedical forgery detection tasks, some limitations remain. First, reliance on modality-specific prompts inferred from a pretrained VLM can lead to suboptimal alignment if the modality is misclassified—particularly for ambiguous or low-quality inputs. Second, although our synthetic dataset captures a broad range of manipulation types, it may not fully represent compositional or multi-step forgeries seen in real-world fraud, such as sequential tampering across figure panels or coordination between image and caption. Lastly, the use of a shared encoder across forgery types may constrain the model’s ability to specialize in task-specific cues. Future work will explore retrieval- or reinforcement-based prompt refinement, validation across a broader set of retracted publications, and reviewer-in-the-loop feedback to support deployment in high-stakes forensic pipelines.

\section{Conclusion}
\label{sec:conclusion}

We introduce a vision-language guided framework for generating and detecting semantic forgeries in biomedical images. By combining diffusion-based synthesis, vision-language prompting, and structured state-space modeling, we construct \textbf{Rescind}—the first large-scale benchmark enabling semantically controlled forgeries across diverse biomedical modalities. To ensure semantic fidelity, we add a vision-language verification loop during generation. We also propose \textbf{Integscan}, a model combining attention-enhanced encoding with prompt-conditioned fusion, achieving state-of-the-art performance across manipulation detection tasks. Code and dataset will be released to support future research in scientific integrity, establishing a scalable foundation for trustworthy and reproducible AI in biomedical forensics.

\section{Ethics Statement}
This work aims to strengthen scientific integrity by improving detection of manipulated biomedical images. All data in Rescind are sourced from publicly available datasets and retracted papers, with no private or patient-identifiable information, and used in accordance with original licenses. To mitigate dual-use risks, all generated images are metadata-tagged, and forgery generation tools will be released only under controlled access for verified research use. We follow responsible stewardship and NIST AI 800-1 guidelines, including license restrictions and rate limits. Malicious actors could theoretically study the benchmark to identify detector weaknesses; however, we believe the benefit of enabling defensive research outweighs this risk. The societal goal is to help journals, integrity offices, and the research community identify fraudulent figures and improve trust in publications. If any party of concern objects to specific images, we will remove them promptly. We acknowledge that synthetic forgeries may not capture the full diversity of real-world misconduct and encourage future work on broader validation.

\bibliography{aaai2026}

\end{document}